\documentclass{llncs}
\usepackage{times}

\begin{document}

\title{A Commentary on ``Breaking Row and Column Symmetries in Matrix Models"}

\author{Alan M. Frisch\inst{1}, Brahim Hnich\inst{2} \and Zeynep Kiziltan\inst{3} \and Ian Miguel\inst{4} \and Toby Walsh\inst{5}}

\institute{University of York, U.K. \\ \email{alan.frisch@york.ac.uk} \and
Monastir University, Tunisia \\ \email{hnich.brahim@gmail.com} 
\and University of Bologna, Italy \\ \email{zeynep.kiziltan@unibo.it} \and University of St Andrews, U.K. \\ \email{ijm@st-andrews.ac.uk}
\and TU Berlin and UNSW Sydney, Germany and Australia \\
\email{tw@cse.unsw.edu.au}}

\newcommand{\SLex}{\mbox{\sc SnakeLex}}
\newcommand{\allperm}{\mbox{\sc AllPerm}}
\newcommand{\DLex}{\mbox{\sc DoubleLex}}

\maketitle

\begin{center}
 \small{30 September 2019}  
\end{center}
 
\begin{abstract}The CP 2002 paper entitled ``Breaking Row and Column Symmetries in Matrix Models" by Flener {\em et al.} \cite{ffhkmpwcp2002} describes some of the first work  for identifying and analyzing row and column symmetry in matrix models and for efficiently and effectively dealing with such symmetry using static symmetry-breaking ordering constraints.  This commentary provides a retrospective on that work and highlights some of the subsequent work on the topic.
\end{abstract}

\section{Setting the scene}

The work described in the CP 2002 publication ``Breaking Row and Column Symmetries in Matrix Models" by Flener {\em et al.} \cite{ffhkmpwcp2002}, henceforth referred to as BRCS, started in early 2001 when Zeynep Kiziltan was a Ph.D. student at Uppsala University (Sweden) and became interested in symmetry breaking in CSPs as a consequence of modelling the rack configuration problem \cite{opl} (prob031 at www.csplib.org).  Zeynep Kiziltan and Brahim Hnich started to investigate alternative models for the problem in order to break more symmetries with respect to the original model. In \cite{opl}, it was argued that  two racks of the same model are equivalent and a permutation of their card assignments  leads to a symmetric solution. This was equivalent to permuting the rows corresponding to the card assignments of the equivalent racks in a matrix of integer decision variables. Without using the term row symmetry (or column symmetry), \cite{opl} proposed to order the first positions of the corresponding rows to prevent the permutations.  Kiziltan and Hnich instead proposed to break this kind of symmetry by ordering the sum of the values in the rows. Both approaches were breaking symmetry via static  ordering constraints. Though useful, they were not able to prevent any possible row permutations in general. This is because two symmetric rows could have the same values in the first positions or  they could have the same sum. Nevertheless, the alternative models and the symmetry breaking constraints were promising for the rack configuration problem, as reported in \cite{rackpaper}. 

\section{Matrix modelling and row and column symmetry}

In May 2001, Alan Frisch, Ian Miguel and Toby Walsh, who were then all at the University of York (U.K.),  visited Uppsala to join Pierre Flener, Justin Pearson, Hnich and Kiziltan to continue a research collaboration which started in late 2000 with Hnich's visit to York. The York group showed interest in  the rack configuration problem  because of its similarity to the steel mill slab design problem (prob038 at www.csplib.org) for which they had recently proposed models \cite{slab}. Both problems can be modelled in a similar way, using matrices of integer decision variables, and their symmetries could be broken likewise, by posting ordering constraints on certain rows. We therefore started to investigate whether there were other problems following this pattern. An obvious example was the famous BIBD problem (prob028 at www.csplib.org).  The symmetries in the standard matrix model of the problem were even more challenging, because any two rows or columns could be permuted, leading to a factorial number of symmetries. One way to tackle such symmetry would be to set the values of the first row and the column of the matrix. While this could help, it would not prevent the permutations of the rows and columns starting with the same value, as in the rack configuration problem. Moreover, all the rows had the same sum, so did all the columns. The sum-based ordering that was used in the rack configuration problem would not work here.  This was when the concept of row and column symmetry, the realization of the factorial number of symmetries in a matrix model with row and column symmetry, and the idea of using a total ordering on vectors of variables to break this type of symmetry started to emerge.  

\section{Breaking row and column symmetry}

A common pattern in CSP models (a matrix of decision variables with two or more dimensions) and an important class of symmetries  arising in such models (row and column symmetry in the 2-dimensional case) were identified. The first total ordering proposed to break this type of symmetry was the lexicographic ordering, such as that used to sort words in a dictionary.  We started to study how to enforce the lexicographic ordering on the symmetric rows and the columns. One possibility was to add  static symmetry-breaking constraints to the model. Another possibility was the so called canonical labelling, which was about compiling the symmetry-breaking constraints directly into the labelling algorithm of search. Both approaches needed attention to ensure that at least one solution in each equivalence class of solutions would remain. We began to address these and related issues over late night dinners and during a fishing trip organized by the hosting department. Another notable outcome of this research visit was the launch of SymCon,\footnote{\url{http://www.it.uu.se/research/group/astra/SymCon/}} the CP workshop series on symmetry in CSPs, which hosted for many years the latest research results in the field.   

The collaboration continued with Kiziltan's return visit to York in September 2001.  Many more problems were shown to be efficiently represented using a matrix model, but again with the challenge of an factorial number of row and column symmetries. It was understood how to consistently add symmetry-breaking lexicographic ordering constraints on the symmetric rows and columns, as well as  how to correctly do canonical labelling. In particular, it was proven that posting simultaneously lexicographic ordering constraints for each symmetric dimension of a matrix model would preserve at least one solution in each equivalence class of solutions. Such constraints are often referred to as $\DLex$ for two dimensions. It was shown that $\DLex$ is generally not complete, in the sense it may leave some symmetries --- in fact, as Gent, Petrie and Puget later showed \cite{handbookSymChapter}, the number of remaining symmetries can be exponential in the size of the matrix. The findings were reported in two papers, \cite{formul01paper} and \cite{symcon01paper}, which would be presented later at the modelling and the symmetry workhops of CP 2001. The only bitter memory of this fruitful collaboration was the attacks on the twin towers and elsewhere on September the 11th, which happened just a few days before the end of Kiziltan's visit to York. Our thoughts go out to the victims and their families.

\section{The initial reaction}

Both papers started to receive interest from the community well before the CP 2001 workhops.  Barbara Smith and Ian Gent began to study breaking row and column symmetry using SBDS, concluding that it can be difficult for SBDS alone to tackle this kind of symmetry \cite{SBDS}. The modelling and the symmetry workshops  provided a constructive platform to discuss the initial findings and the challenges in breaking row and column symmetry in matrix models, laying the foundations for BRCS as well as the research on the topic for the next 15 years. 

After BRCS was published, we learnt that the first results on constraints for  breaking symmetry on multiple dimensions of a matrix came from Lubiw \cite{lubiw}, who showed that $\DLex$ is a consistent symmetry-breaking constraint for a two-dimensional matrix that has row and column symmetry.  Unaware of Libuw's work, Shlyakhter \cite{ilya01} and we independently generalised Libuw's result to multiple dimensions. \footnote{Actually, Lubiw and Shlyakhter each considered only the
case of matrices containing zeros and ones, but their proofs
trivially generalise to any set of values.} 
Our and Lubiw's proofs show that any matrix that does
not satisfy the constraint can be permuted into one that does.
In contrast, Shlyakhter's proof shows that $\DLex$ is entailed by the row-wise lex-leader constraints for the
symmetry group \cite{crawford}.  By the row-wise lex-leader constraints we mean those obtained by using the lex-leader schema on the vector of variables obtained by taking the first row of the vector, followed by the second row, and  so forth until all rows are included.


The identification of lexicographic ordering as a means of dealing with row and column symmetry inspired research into propagating this constraint effectively. At the same conference at which BRCS was published, Frisch et al. \cite{frisch2002global} introduced an efficient propagator for establishing generalised arc consistency on a lexicographic ordering constraint between a pair of vectors of variables  (see also the later journal paper \cite{frisch2006propagation}). 

\section{Subsequent work}

The incompleteness of $\DLex$, the difficulty of breaking all row and column symmetry in general, and the need of propagating $\DLex$ efficiently opened up several research directions. While it is not possible to mention them all in this commentary, we would like to highlight some of the subsequent work in the context of static symmetry breaking. Let us consider two-dimensional matrices. One line of research focused on tractable cases. Katsirelos  {\em et al.} proved that in certain special cases all row and column symmetry can be broken with a polynomial number of constraints  \cite{ffhkmpwcp2002}, the associated symmetry breaking  constraints can be propagated in polynomial time \cite{cp2010}. They also showed that in a matrix with a bounded number of rows (or columns),  all row and column symmetry can be broken in polynomial time. Yip and Van Hentenryck \cite{pascalijcai11} turned this theoretical result into a complete and efficient method for breaking all row and column symmetry in matrix models with a small number of rows (or columns). 

Following \cite{frisch2006propagation}, Carlsson and Beldiceanu developed a propagator for a chain of lexicographic ordering constraints, such as for the successive rows or columns of a matrix \cite{carlsson2002arc}. Many efficient and simple decompositions of the lexicographic ordering constraint on a pair of vectors of variables followed up, such as those reported in \cite{constraintchecker} and \cite{slide}.  

Another line of research looked into other constraints that can be effectively added in conjunction with $\DLex$ to break more row and column symmetry.  Frisch and Harvey \cite{frisch:2003} considered the complete row-wise lex-leader constraints for row and column  symmetry of a matrix of two rows and three columns.  They simplified these constraints and noticed that they contained constraints asserting that the first row is lexicographically less or equal to \emph{every} permutation of the second row.  This observation led to the work of Frisch {\em et al.}  \cite{allperm}, who showed that generally the row-wise lex-leader constraints entail that the first row is lexicographically less than or equal to every permutation of every other row. They called the constraint ensuring this property $\allperm$, introduced a propagator for it, and obtained better results than $\DLex$ alone when posted together with $\DLex$.  

Another interesting direction has been the investigation alternatives to, rather than extensions of $\DLex$.  Promising results have been obtained by $\SLex$ constraints, which derive from lex-leader constraints obtained by ordering the matrix variables not row-wise but by snaking through the matrix \cite{slex}. Alternatives to lexicographic ordering have been studied, including  multiset ordering \cite{multiset,frisch2009filtering}, Gray code ordering \cite{gray} and reflex ordering \cite{reflex}.  

\section{Conclusions} 
We are pleased to see that following the publication of the initial CP 2001 workshop papers and BRCS, breaking row and column symmetries in matrix models has been an active research area leading to many publications and the common inclusion of the global lexicographic ordering constraints on vectors in constraint modelling languages and solvers such as Choco, OPL, MiniZinc, Sicstus Prolog, Gecode. This commentary hopefully makes it clear that it is time we produced a survey on this important modelling and solving topic.


\begin{thebibliography}{10}
\providecommand{\url}[1]{\texttt{#1}}
\providecommand{\urlprefix}{URL }
\providecommand{\doi}[1]{https://doi.org/#1}

\bibitem{constraintchecker}
Beldiceanu, N., Carlsson, M., Petit, T.: Deriving filtering algorithms from
  constraint checkers. In: Proc. of the 10th International Conference on the
  Principles and Practices of Constraint Programming (CP-2004). Springer (2004)

\bibitem{slide}
Bessiere, C., Hebrard, E., Hnich, B., Kiziltan, Z., Walsh, T.: Slide: A useful
  special case of the {\sc cardpath} constraint. In: Proc. of the 18th European
  Conference on Artificial Intelligence (ECAI-2008). IOS Press (2008)

\bibitem{carlsson2002arc}
Carlsson, M., Beldiceanu, N.: Arc-consistency for a chain of lexicographic
  ordering constraints. Tech. rep., Swedish Institute of Computer Science
  (November 2002)

\bibitem{crawford}
Crawford, J., Ginsberg, M., Luks, E., Roy, A.: Symmetry breaking predicates for
  search problems. In: Proc. of the 5th International Conference on the
  Principles of Knowledge Representation and Reasoning (KR-1996). Morgon
  Kaufmann (1996)

\bibitem{symcon01paper}
Flener, P., Frisch, A.M., Hnich, B., Kiziltan, Z., Miguel, I., Pearson, J.,
  Walsh, T.: Symmetry in matrix models. In: Proc. of SymConÕ01, the CPÕ01
  Workshop on Symmetry in Constraint Satisfaction Problems (2001)

\bibitem{ffhkmpwcp2002}
Flener, P., Frisch, A.M., Hnich, B., Kiziltan, Z., Miguel, I., Pearson, J.,
  Walsh, T.: Breaking row and column symmetry in matrix models. In: Proc. of
  the 8th International Conference on the Principles and Practices of
  Constraint Programming (CP-2002). Springer (2002)

\bibitem{formul01paper}
Flener, P., Frisch, A.M., Hnich, B., Kiziltan, Z., Miguel, I., Walsh, T.:
  Matrix modelling. In: Proc. of FormulÕ01, the CPÕ01 Workshop on Modelling
  and Problem Formulation, (2001)

\bibitem{frisch:2003}
Frisch, A.M., Harvey, W.: {Constraints for breaking all row and column
  symmetries in a three-by-two-matrix}. In: Proc. of SymConÕ03, the CPÕ03
  Workshop on Symmetry in Constraint Satisfaction Problems (2003)

\bibitem{frisch2002global}
Frisch, A., Hnich, B., Kiziltan, Z., Miguel, I., Walsh, T.: Global constraints
  for lexicographic orderings. In: Proc. of the 8th International Conference on
  the Principles and Practices of Constraint Programming (CP-2002. Springer
  (2002)

\bibitem{frisch2006propagation}
Frisch, A.M., Hnich, B., Kiziltan, Z., Miguel, I., Walsh, T.: Propagation
  algorithms for lexicographic ordering constraints. Artificial Intelligence
  \textbf{170}(10),  803--834 (2006)

\bibitem{frisch2009filtering}
Frisch, A.M., Hnich, B., Kiziltan, Z., Miguel, I., Walsh, T.: Filtering
  algorithms for the multiset ordering constraint. Artificial Intelligence
  \textbf{173}(2),  299--328 (2009)

\bibitem{multiset}
Frisch, A., Hnich, B., Kiziltan, Z., Miguel, I., Walsh, T.: Multiset ordering
  constraints. In: Proc. of the 18th International Joint Conference on
  Artificial Intelligence (IJCAI-2003). Morgan Kaufmann (2003)

\bibitem{allperm}
Frisch, A., Jefferson, C., Miguel, I.: Constraints for breaking more row and
  column symmetries. In: Proc. of the 9th International Conference on the
  Principles and Practice of Constraint Programming (CP-2003). Springer (2003)

\bibitem{slab}
Frisch, A., Miguel, I., Walsh, T.: Modelling a steel mill slab design problem.
  In: Proc. of the IJCAIÕ01 Workshop on Modelling and Solving Problems with
  Constraints (2001)

\bibitem{handbookSymChapter}
Gent, I., Petrie, K., Puget, J.F.: Symmetry in constraint programming. In:
  Handbook of Constraint Programming, pp. 329--376 (2006)

\bibitem{slex}
Grayland, A., Miguel, I., Roney-Dougal, C.: Snake lex: An alternative to double
  lex. In: Proc. of the 15th International Conference on the Principles and
  Practice of Constraint Programming (CP-2009). Springer (2009)

\bibitem{cp2010}
Katsirelos, G., Narodytska, N., Walsh, T.: On the complexity and completeness
  of static constraints for breaking row and column symmetry. In: Proc. of the
  16th International Conference on the Principles and Practice of Constraint
  Programming (CP-2010). Springer (2010)

\bibitem{rackpaper}
Kiziltan, Z., Hnich, B.: Symmetry breaking in a rack configuration problem. In:
  Proc. of the IJCAIÕ01 Workshop on Modelling and Solving Problems with
  Constraints (2001)

\bibitem{reflex}
Lee, J.H.M., Zhu, Z.: Static symmetry breaking with the reflex ordering. In:
  Proc. of the 25th International Joint Conference on Artificial Intelligence
  (IJCAI-2016). IJCAI/AAAI Press (2016)

\bibitem{lubiw}
Lubiw, A.: Doubly lexical orderings of matrices. In: Proc. of the 17th Annual
  Association for Computing Machinery Symposium on Theory of Computing
  (STOC-85). ACM Press (1985)

\bibitem{gray}
Narodytska, N., Walsh, T.: Breaking symmetry with different orderings. In:
  Proc. of the 19th International Conference on the Principles and Practice of
  Constraint Programming (CP-2013). Springer (2013)

\bibitem{ilya01}
Shlyakhter, I.: Generating effective symmetry-breaking predicates for search
  problems. Electronic Notes in Discrete Mathematics  \textbf{9},  19--35
  (2001)

\bibitem{SBDS}
Smith, B.M., Gent, I.P.: Reducing symmetry in matrix models: Sbds vs.
  constraints. In: Proc. of SymConÕ01, the CPÕ01 Workshop on Symmetry in
  Constraint Satisfaction Problems (2001)

\bibitem{opl}
{van Hentenryck}, P.: The OPL Optimization Programming Language. The MIT Press
  (1999)

\bibitem{pascalijcai11}
Yip, J., {van Hentenryck}, P.: Symmetry breaking via lexleader feasibility
  checkers. In: Proc. of the 22nd International Joint Conference on Artificial
  Intelligence (IJCAI-2011). IJCAI/AAAI (2011)

\end{thebibliography}

\end{document}